\documentclass[11pt,a4paper]{article} 
\usepackage[a4paper,margin=1in]{geometry}
\usepackage{times}
\usepackage[utf8]{inputenc} 
\usepackage[T1]{fontenc}    
\usepackage{url}            
\usepackage{booktabs}       
\usepackage{amsfonts}       
\usepackage{nicefrac}       
\usepackage{microtype}      
\usepackage{xcolor}         

\usepackage{graphicx}
\usepackage{booktabs} 

\usepackage{tcolorbox} 
\usepackage{lipsum} 
\usepackage{paralist}	
\usepackage{tabularx}
\usepackage{threeparttable}
\usepackage{makecell}
\usepackage{multirow}
\usepackage{xspace}

\usepackage{pifont}

\usepackage{subcaption}
\usepackage[numbers,sort&compress]{natbib}
\usepackage{amsmath}
\usepackage{amssymb}
\usepackage{mathtools}
\usepackage{amsthm}
\usepackage[normalem]{ulem} 
\usepackage[textsize=tiny]{todonotes}
\usepackage{siunitx} 
\usepackage{hyperref} 
\usepackage[capitalize,noabbrev]{cleveref}
\theoremstyle{plain}

\theoremstyle{definition}

\theoremstyle{remark}

\newcolumntype{S}{>{\hsize=3.5\hsize\raggedright\arraybackslash}X}
\newcolumntype{N}{>{\hsize=0.75\hsize\centering\arraybackslash}X}

\newcommand{\claims}{ClaimRAG-LAW\xspace}

\title{Fine-grained Claim-level RAG Benchmark for Law}

\author{
	Souvick Das \and Sallam Abualhaija \and Domenico Bianculli\\
	SnT, University of Luxembourg\\
	Luxembourg\\
	\texttt{\{souvick.das,sallam.abualhaija,domenico.bianculli\}@uni.lu}
}

\begin{document}

\maketitle

\begin{abstract}

The rapid progress of large language models (LLMs) is shifting 
semantic search toward a question-answering paradigm, 
where users ask questions and LLMs generate responses. In  
high-stake domains such as law, retrieval-augmented  
generation (RAG) is commonly used to mitigate 
\textit{hallucinations} in generated responses. 
Nonetheless, prior work shows that RAG systems, 
whether general-purpose or legal-specific, still hallucinate 
at varying rates,
making fine-grained evaluation essential.
Despite the need, existing evaluation frameworks for legal RAG 
systems lack the granularity
required to provide detailed analysis of retrieval and 
generation performance separately. 
Moreover, current benchmarks are largely English-only and 
centered on legal expert queries, overlooking non-expert 
needs.
We introduce \claims, a comprehensive dataset for legal RAG 
that supports French and English, targets both experts and 
non-experts, and includes diverse question types reflecting 
realistic scenarios. We further apply a fine-grained evaluation
framework
of state-of-the-art legal RAG systems, revealing limitations in 
retrieval, generation, and claim-level analysis in the legal domain. 

\end{abstract}

 \section{Introduction}
\label{sec:introduction}

The rapid progress of large language models (LLMs) has 
revolutionized information access, with growing reliance on 
chatbots shifting semantic search toward a 
question-answering (QA) paradigm in which users pose 
questions and LLMs generate 
responses~\cite{alammar2024}. 
This shift can be critical in high-stakes domains such as law, where
adoption of LLMs among legal professionals is growing at a 
rapid pace~\cite{edwards2025}. 
Retrieval-augmented generation (RAG)~\cite{lewis2020} 
grounds LLM outputs in authoritative sources, increasing 
response reliability~\cite{keisha2025}. 
Despite their benefits, their use in the legal domain remains 
problematic, even with RAG, due to persistent hallucinations 
(i.e., 
generating incorrect information~\cite{dahl2024}) and a 
tendency to be perceived as more reliable than they actually 
are~\cite{mallick2024,magesh2025}. 

Theoretical work shows that LLMs must hallucinate to a 
certain extent, irrespective of their architecture, training data 
quality, or scale~\cite{kalai2024}. 
Empirical evidence reinforces this limitation in the legal domain:
\citet{dahl2024} show that general-purpose LLMs hallucinate 
in approximately 58\% to 82\% of cases when applied to legal 
queries.  
\citet{magesh2025} report hallucination rates of 17\% to 33\% 
even for customized legal AI systems, challenging claims that 
legal research tools are substantially less prone to 
hallucination. 
Real-world incidents further emphasize these findings, from 
early ChatGPT-related failures 
in legal practice~\cite{weiser2023} to court sanctions resulting 
from AI-generated legal filings~\cite{esquire2025}. 

The prevalence of hallucinations in both general-purpose and 
legal-specific LLMs, even with RAG, highlights the need for 
dedicated legal evaluation benchmarks. 
Early legal benchmarks such as LEXTREME~\cite{niklaus2023} 
and 
LegalBench~\cite{guha2023} 
established important baselines for legal natural language 
processing (NLP), but were not designed to assess RAG 
systems, which now dominate the landscape of semantic 
search and QA.

Legal RAG benchmarks remain limited. 
LegalBench-RAG~\cite{pipitone2024} 
focuses on a narrow 
set of
legal retrieval scenarios, 
while the benchmark of 
\citet{magesh2025} is specific to U.S. law and relies on costly 
manual evaluation. 
More broadly, existing benchmarks 
often lack the fine-grained analysis of retrieval and 
generation, are predominantly English-centric, 
and primarily target legal professionals rather than 
supporting the general public in accessing legal information.  

To enable fine-grained evaluation of RAG systems,  
\citet{trulens} introduced the RAG Triad,  which 
assesses context relevance, groundedness, and answer 
relevance. 
This framework has since inspired a variety of claim-based 
evaluation approaches~\citet{ru2024,li2025}, 
where a claim is
defined as a minimal, self-contained factual 
statement~\cite{metropolitansky2025}.
Despite the effectiveness in 
hallucination detection~\cite{scire2024}, 
claim-based approaches 
remain largely 
under-explored in the legal domain.

This
paper makes three key contributions:
\begin{compactenum}
\item We introduce a bilingual benchmark \claims 
(``Fine-grained Claim-level RAG Benchmark for 
Law'') for fine-grained evaluation of legal RAG. 
The dataset and its generation code are publicly 
available\footnote{Links omitted for double blind.}. %
Designed to be \textit{diverse},  
  \textit{representative}, 
and  \textit{fine-grained}, 
  \claims covers multiple 
  question
  categories (adopted from \citet{magesh2025}) and 
  user personas, from legal
  professionals to lay users
  seeking access to legal 
  information. It comprises a total of 317 QA pairs sourced from 
  the general 
  data   protection  regulation (GDPR)~\cite{EUGDPR} (English) 
  and the Luxembourg national 
  civil law~\cite{CIVIL} (French)
  (hereafter referred to as \textit{CIVIL}). Additionally, it 
  contains 968 manually-validated claims to support 
  claim-level evaluation of legal RAG systems. 

\item We adopt fine-grained metrics  proposed 
by~\citet{ru2024} and assess their applicability to legal RAG 
systems.

\item We benchmark 
state-of-the-art RAG systems on \claims, uncovering  
domain-specific limitations in retrieval and generation. 
We further evaluate 
claim extraction and analysis in legal settings, revealing 
important limitations, particularly in detecting contradictory  
claims. 
  
\end{compactenum}

The remainder of the paper is structured as follows. 
Section~\ref{sec:related} surveys related work. 
Section~\ref{sec:ClaimRAG-Law} details the construction 
of \claims. Section~\ref{sec:evaluation} reports the results of 
our empirical evaluation. Section~\ref{sec:conclusion} 
concludes the paper.

 \section{State of the art}
\label{sec:related}

This section reviews two 
research directions: (1) legal RAG benchmarks, 
and (2) claim-level evaluation for
hallucination detection. 
We identify two key research gaps that motivate our work: 
(1) existing legal RAG benchmarks lack 
comprehensive evaluation compared to general domains; and 
(2) fine-grained, claim-level evaluation remains largely 
under-explored in the legal domain despite its success in 
general-domain RAG. 

\subsection{RAG Benchmarks in Legal Domain}
Early legal benchmarks, including
LexGLUE~\cite{chalkidis2022},   
LEXTREME~\cite{niklaus2023},  
and LegalBench~\cite{guha2023} significantly advanced the 
evaluation of 
legal NLP and LLM reasoning across diverse tasks and 
languages. 
However, these benchmarks primarily focus on classification, 
prediction, and reasoning tasks rather than 
 RAG. As a result, they are not well suited for assessing 
 retrieval-grounded responses or measuring hallucinations in 
 legal 
 applications.

More recently, dedicated benchmarks for legal RAG 
evaluation have emerged. 
CLERC~\cite{hou2025} 
provides a large‑scale 
dataset for case-law retrieval, while 
LegalBench-RAG~\cite{pipitone2024}, which extends  
LegalBench~\cite{guha2023}, measures the retrieval 
precision for contract‑related questions. Despite their 
contributions, both benchmarks
are primarily retrieval-centric and do not comprehensively 
assess 
end-to-end answer generation.

To address this limitation, 
LLeQA~\cite{louis2024} introduced an expert-annotated  
French benchmark for long-form statutory QA based on 
Belgian law, 
enabling evaluation of retrieve-then-read RAG systems. 
Nevertheless, its scope remains limited to a single language 
and 
jurisdiction. Moreover, it 
relies largely on holistic automatic metrics and does not 
provide 
fine-grained evaluation of generated answers.

Several other legal RAG benchmarks have been proposed for 
specific legal systems, 
LexRAG~\cite{li2025} evaluates 
legal consultation dialogues in Chinese law, 
CBR-RAG~\cite{wiratunga2024}  focuses on case-based 
retrieval in Australian law, and 
KOBLEX~\cite{lee2025koblex} targets provision-grounded 
multi-hop QA in Korean law. 
While these benchmarks contribute valuable 
jurisdiction-specific resources, their focus on individual legal 
systems limits the ability to draw cross-jurisdictional 
conclusions.

Overall, existing legal RAG benchmarks provide only partial 
coverage of the challenges involved in retrieval-grounded 
legal question answering. Although 
LLeQA partially addresses non-expert information needs, no 
benchmark systematically evaluates RAG systems from a 
non-expert perspective across multiple jurisdictions and 
languages. Furthermore, current datasets offer limited support 
for fine-grained analysis that jointly examines retrieval quality, 
answer generation, and hallucination behavior. 

In parallel, the broader RAG community has developed 
evaluation frameworks such as RAGAS~\cite{es2024} and 
RAGChecker~\cite{ru2024}, enabling more detailed 
assessment of retrieval and generation performance. In 
contrast, equivalent benchmarking resources for legal 
retrieval-grounded generation remain relatively scarce and are 
still in an early stage of development.

\subsection{Claim-Level Evaluation}
The rise of generative AI in law has exposed the limitations of 
traditional evaluation metrics such as ROUGE and BLEU, which 
measure lexical overlap but often fail to capture factual 
correctness and hallucinations in RAG systems~\cite{ru2024}. 
To address this gap, frameworks such as 
TruLens~\cite{trulens} introduced the RAG Triad, which 
evaluates context relevance, groundedness, and answer 
relevance. Although influential in general-domain RAG 
evaluation, these approaches have seen limited adoption in 
legal applications.

This limitation is particularly important in law, where answers 
may closely resemble reference texts while containing subtle 
factual errors. As noted in LLeQA~\cite{louis2024}, high 
semantic similarity does not guarantee factual correctness, 
making fine-grained, claim-level evaluation essential.

Several frameworks have therefore been proposed to assess 
factuality at the claim level. FActScore~\cite{min2023} 
decomposes responses into atomic facts and measures the 
proportion supported by external evidence. 
RefChecker~\cite{hu2024} extracts subject-predicate-object 
triples and uses entailment-based reasoning to identify 
unsupported claims. FENICE~\cite{scire2024} combines claim 
extraction with natural language inference (NLI) to provide 
interpretable, multi-granular factuality assessments, while 
SAFE~\cite{wei2024} verifies extracted facts against web 
evidence. More recently, 
Claimify~\cite{metropolitansky2025_clamify} highlighted the 
importance of claim extraction itself, showing that inaccurate 
or incomplete claims can lead to misleading hallucination 
assessments.

Despite these advances, hallucination detection remains an 
open 
challenge~\cite{manakul2023,wang2025,metropolitansky2025}.
 Legal LLMs and RAG systems continue to generate 
unsupported or misleading 
statements~\cite{dahl2024,magesh2025}. An empirical study 
of RAG-based legal research tools found hallucination rates 
between 17\% and 33\%, with \emph{misgrounding} as the dominant 
failure mode: systems often cite genuine legal sources that do 
not actually support the generated 
answer~\cite{magesh2025}. Such errors are unlikely to be 
captured by holistic evaluation metrics, underscoring the need 
for claim-level analysis in legal settings. 
However, 
legal-specific methods remain scarce. RePASs (Regulatory 
Passage Answer Stability Score)~\cite{gokhan2024} 
introduces an NLI-based metric for evaluating contradiction 
and obligation coverage in regulatory texts, but its scope is 
limited to Gulf financial regulations. Consequently, a gap 
remains for fine-grained, claim-level evaluation of multilingual 
statutory-law RAG systems.


\section{ClaimRAG-Law}
\label{sec:ClaimRAG-Law}

We present \textit{\claims}, a  
benchmark for evaluating legal RAG systems. It consists of 
two complementary 
datasets: 
one for end-to-end legal RAG evaluation and another 
for assessing claim-checking methods. 

\subsection{Desiderata} 
\label{subsec:desiderata}

The design of \claims is guided by three desiderata derived 
from the gaps identified in the literature.

\paragraph{(1) Diverse. } 
Legal RAG systems serve users with varying levels of expertise 
and information needs. For example, some users seek general 
legal guidance, others require precise factual references, while 
some may pose questions based on incorrect assumptions. 
To capture this diversity, \claims includes QA pairs spanning 
four question categories inspired by~\citet{magesh2025}:

(1) \textit{General legal research questions} represent a 
typical 
use case for legal RAG systems, where users ask general 
questions about, e.g., common-law doctrines, specific 
holdings of court cases, or interpretations of regulations.

(2) \textit{Factual recall questions} target verifiable details that
require minimal legal interpretation, such as citations, effective
dates of legislation, or specific entities mentioned in a
clause. 

(3) \textit{False premise questions} begin from an incorrect 
assumption about a legal fact and  therefore require the 
system to identify and correct the erroneous premise. 

(4) \textit{Jurisdiction or time-specific questions} capture  the 
challenge that laws vary across jurisdictions and evolve over 
time.

These questions are designed to represent the needs of legal 
experts, public officials with partial legal knowledge, and 
laypersons with limited ability to verify generated answers.

\paragraph{(2) Representative. } Evaluating whether RAG 
systems generate legally grounded answers, as opposed to 
relying on parametric knowledge, requires coverage across 
multiple jurisdictions and languages. To this end, 
\claims draws its QA pairs from two complementary legal 
sources: 
(1) the General Data Protection Regulation 
(GDPR)~\cite{EUGDPR}, a supranational European regulatory 
framework (English version), and (2) a national Civil Code 
(CIVIL)~\cite{CIVIL} (French version). These sources 
provide variation in both jurisdictional scope and language, 
enabling a more representative evaluation of legal RAG 
systems.

\paragraph{ (3) Fine-grained. } Recent RAG evaluation 
research increasingly relies on claim-level analysis to assess 
factual grounding and hallucinations. However, most 
approaches assume that claim extraction and verification are 
sufficiently reliable. To support fine-grained legal RAG 
evaluation, \claims includes a claim-level dataset derived from 
approximately half of the QA pairs. Claims were automatically 
extracted and subsequently validated by a legal expert, who 
also annotated their logical relation to the supporting legal 
context. This enables detailed analysis of retrieval quality, 
factual grounding, and hallucination behavior in generated 
legal answers.

\subsection{The dataset} \label{subsec:statistics}

Following the design principles outlined above, we constructed 
\claims using a two-stage methodology. First, QA pairs were 
generated automatically from the source regulations. Second, 
all generated QA pairs were manually reviewed and validated 
by a legal expert. Additional details on the generation and 
validation procedures are provided in 
Appendix~\ref{sec:appendix}. 

\begin{figure*}[t] \centering 
	\includegraphics[width=.8\linewidth]{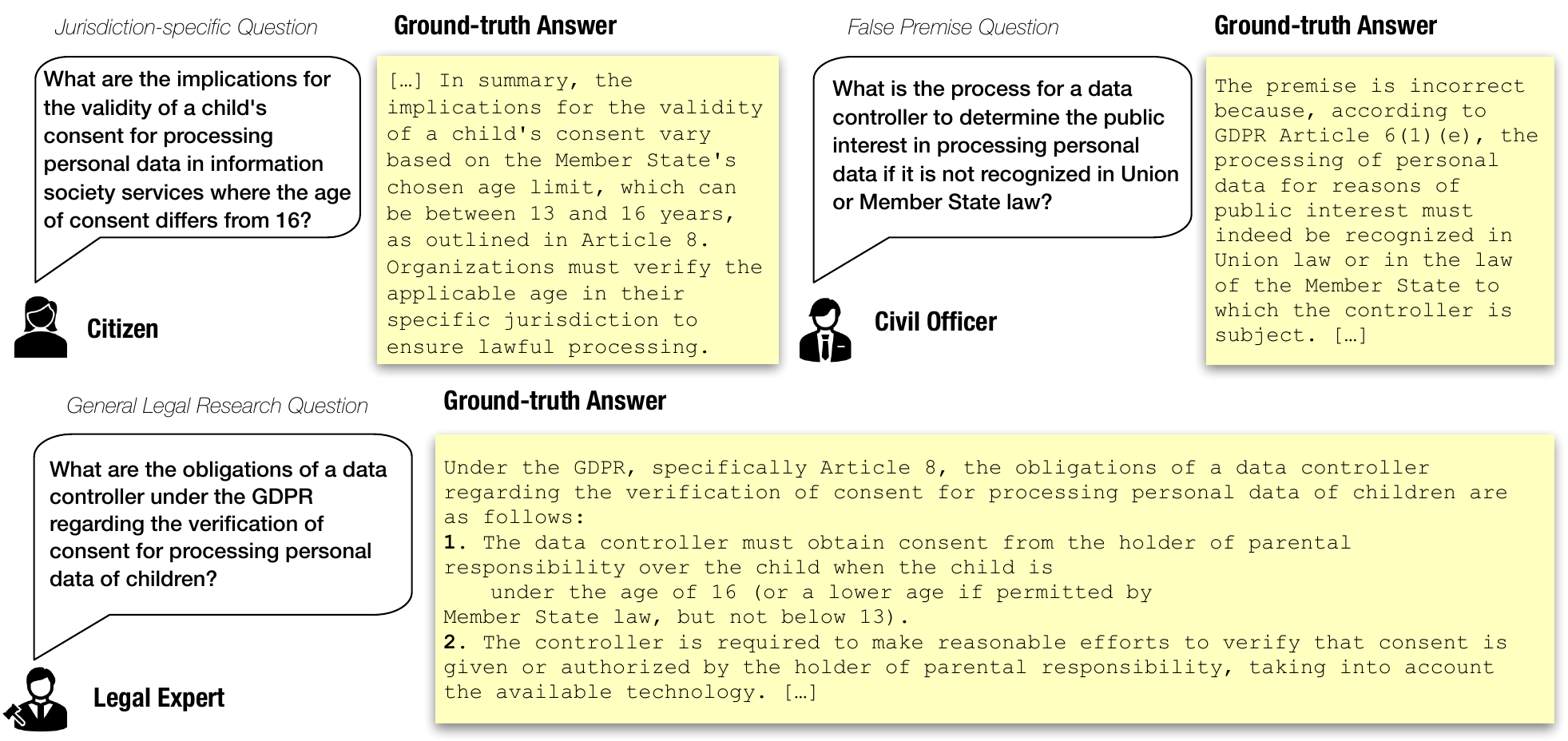} 
	\caption{Example of GDPR questions and ground-truth 
	answers in 
		\claims.} 
	\label{fig:question_examples} 
\end{figure*}

\claims comprises 317 legal 
QA pairs drawn from 
two complementary legal sources: GDPR (English) and CIVIL 
(French). Table~\ref{tab:combined_stats} 
summarizes the dataset statistics. 
Fig.~\ref{fig:question_examples} shows example questions 
from different categories together with their corresponding 
ground-truth answers from \claims (GDPR). More examples 
are 
provided in Section~\ref{sec:appendix-b}.

To support fine-grained evaluation, we additionally 
constructed a claim-level dataset from a subset of the QA 
pairs. In total, \claims contains 968 manually validated claims, 
of which 86.6\% (838/968) were judged by the legal expert to 
be correct with respect to the source legal context. 
This claim-level annotation enables detailed analysis of factual 
grounding and hallucinations in legal RAG systems.

\begin{table*}[t]
	\centering
	\footnotesize
	\caption{Statistics for \claims Dataset 
	}
	\label{tab:combined_stats}

		\begin{threeparttable} 

	\begin{tabularx}{\textwidth}{
					l
					c
					*{3}{>{\centering\arraybackslash}X}
					*{4}{>{\centering\arraybackslash}X} | 
					*{6}{>{\centering\arraybackslash}X}
				}
			\toprule
			&  
			& \multicolumn{3}{c}{\textbf{Persona}} 
			& \multicolumn{4}{c}{\textbf{Question Category}} 
			& \multicolumn{6}{c}{\textbf{Claims}} \\
			\cmidrule(lr){3-5} \cmidrule(lr){6-9} 
			\cmidrule(lr){10-15}
			
			\textbf{} 
			& $\sum_\text{QA}$ 
			& P$_1$ &  P$_2$ &  P$_3$
			& CG$_1$ & CG$_2$ & CG$_3$ & CG$_4$
					& $\sum_\text{A}$ 
					& $\sum_\text{C}$  & Valid & $\epsilon$ & 
					$\eta$ 
					& 
					$\zeta$ \\
			\midrule
			
			\textbf{GDPR (EN) }
			& 149 
			& 20 & 77 & 52 
			& 131 & 9 & 8 & 1 
			& 63
			& 520 & 453 & 295 & 151 & 7 \\
			
			\textbf{CIVIL (FR)  }
			& 168 
			& 35 & 29 & 104 
			& 137 & 24 & 5 & 2 
			& 89
			& 448 & 385 & 284 & 77 & 24 \\
			\midrule
			
			\textbf{Total} 
			& \textbf{317}
			& \textbf{55} & \textbf{106} & \textbf{156} 
			& \textbf{268} & \textbf{33} & \textbf{13} & \textbf{3} 
			& \textbf{152}
			& \textbf{968} & \textbf{838} & \textbf{579} & \textbf{228} & \textbf{31} \\
			\bottomrule
		\end{tabularx}
	    \begin{tablenotes} 
		\small
		\it \item[1] $\sum_\text{QA}$ refers to the number of QA 
		pairs. 
		P$_1$, P$_2$, P$_3$ refer to the personas: citizen, 
		civil officer, and legal 
		expert, respectively. CG$_1$, CG$_2$, CG$_3$, 
		CG$_4$ 
		refer 
		to the question categories: 
		general legal research, factual recall, false premise, 		
		jurisdiction/time-specific, respectively. 
		\item[2] $\sum_\text{A}$: the number of generated 
		answers 
		used to 
		generate the claims, and $\sum_\text{C}$: the number 
		of 
		claims. Valid indicates the number of claims deemed 
		correct 
		by the expert, and $\epsilon$, $\eta$, $\zeta$ refer to 
		entailment, neutral, 
		and 
		contradiction, respectively.

	\end{tablenotes}
	\vspace*{-1.5em}
	\end{threeparttable}

\end{table*}


\section{Benchmarking RAG Systems in Legal Domain}
\label{sec:evaluation}

\subsection{Research Questions}
\label{sec:rqs}

\textbf{RQ1. How well do existing RAG systems 
perform in legal settings? }
This RQ provides a fine-grained assessment of 
RAG systems in
legal settings by assessing both their overall performance on 
our curated dataset and the performance of their retrieval and 
generator components separately.
To achieve this, we use \textit{RAGChecker}~\cite{ru2024}, a
fine-grained RAG evaluation framework. We selected 
RAGChecker for two reasons. First, it has shown to provide a
detailed view of the performance of  RAG systems. Second,
\citet{ru2024} report that its metrics correlate more strongly 
with human judgments in terms of correctness (both Pearson 
and Spearman) compared to 
alternative frameworks such as
RAGAs~\cite{es2024}. This alignment with human evaluation is 
particularly important in the legal domain, where subtle 
differences may not be captured by aggregated
metrics.

\textbf{RQ2. How accurate are automated claim
extraction and verification methods in the legal domain?}
Since claims form the basis of fine-grained RAG evaluation, 
this RQ assesses the reliability of 
claim-level analysis in the legal domain.  Specifically, 
we evaluate the accuracy of RefChecker~\cite{hu2024}  in 
extracting legal claims 
from the generated 
responses and verifying their entailment relationship with 
a reference text. 
We focus on RefChecker because it serves as the 
claim-level analysis engine underlying RAGChecker, which is 
used in RQ1.

\subsection{Accuracy of RAG Systems in Legal 
Settings 
(RQ1)}
\label{sec:rq1}

To answer RQ1, we evaluated the performance of several 
RAG systems on \claims using the fine-grained evaluation 
framework proposed by~\citet{ru2024}.

\noindent\textit{\textbf{RAG  Systems. }}
We evaluated eight RAG systems obtained by combining two 
retrieval methods with four LLM-based generators. 
For retrieval, we used the same retrievers considered 
by~\citet{ru2024}: 
\textit{BM25}~\cite{robertson2009}, a standard sparse 
retriever, and
\textit{E5-Mistral-7B-Instruct}~\cite{wang2023}, a
state-of-the-art dense retriever.
For generation, we adopted three of the models used 
by~\citet{ru2024}:  
\textit{GPT-4o}~\cite{hurst2024},
\textit{Llama-3.1-8B-Instruct}~\cite{grattafiori2024}, and 
\textit{Mixtral-8x7B}~\cite{jiang2024}. 
We excluded \textit{Llama-3.1-70B-Instruct} because of its high 
computational cost. In addition, we included 
GPT5~\cite{singh2025}  to assess whether a newer generation 
of the same model family provides advantages in legal 
QA.

\noindent\textit{\textbf{Implementation Details. }}
Following \citet{ru2024}, we implemented both retrievers 
using 
OpenSearch index (\url{https://opensearch.org/}) and 
E5-Mistral-7B-Instruct tokenizer for chunking. 
We split documents into fixed-size chunks of 600 
tokens with an overlap of 120 tokens. 
We prefer fixed-size chunking over 
article-level chunking because legal documents, such 
as GDPR and CIVIL, exhibit substantial variation in article 
length, resulting in inconsistent retrieval granularity. 
We chose 600-token chunks based on the ablation study of 
~\citet{ru2024}, which showed consistent improvements in 
claim recall and F1 as chunk size increases. 

We retrieved the top-$k=10$, corresponding to 6000 tokens of 
retrieved context. Together with  
the system prompt, question, 
and generated response, this remained within the 
8192-token context window of Llama-3.1-8B-Instruct, the 
most constrained generator in our study. 
For generation, we used the default prompt template 
of~\citet{ru2024}, a temperature of 0.0, and a single 
evaluation run.   
We conducted all experiments  on a workstation with a single 
NVIDIA RTX 3090 GPU 
(\SI{24}{\giga\byte}), an AMD Ryzen 9 5900X 12-core CPU, 
and \SI{64}{\giga\byte} RAM.

\noindent\textit{\textbf{RAGChecker Input Construction. }}
RAGChecker requires a tuple consisting of a 
question,
retrieved chunk (context), generated response, and  
ground-truth answer. In \claims, the
question and ground-truth answer are provided, whereas the 
retrieved chunk and generated response are 
produced by each RAG system. 
For every question, the system retrieves the top-$k$ relevant 
chunks from the corresponding legal document (GDPR or 
CIVIL) and generates a response.  
The resulting tuples are then evaluated with RAGChecker. 

\noindent\textit{\textbf{Metrics. }}
We use the fine-grained metrics proposed by \citet{ru2024}. 
These metrics are based on claim extraction and entailment 
analysis between claims and reference texts. 

\textit{Overall metrics} include (i) \textit{precision (P)}, the
proportion of response claims entailed by
the ground-truth answer; (ii) \textit{recall (R)}, the proportion 
of ground-truth claims correctly identified by the model 
response; and \textit{F1 score}, the harmonic 
mean between P and R. 

\textit{Retriever metrics}  include 
(i) \textit{claim recall (CR)}, the proportion of ground-truth 
claims entailed by the retrieved chunks; and
(ii) \textit{context precision (CP)}, the proportion of retrieved 
chunks entailing at least one ground-truth claim. 

\textit{Generator metrics} include 
(i) \textit{faithfulness (FT)}, the proportion of response 
claims entailed by the retrieved chunks;  
(ii) \textit{self knowledge (SK)}, the proportion of correct 
response claims not entailed by the retrieved context; 
(iii) \textit{context utilization (CU)}, the ratio of correct 
response claims supported by the retrieved chunks to the total 
relevant ground-truth claims entailed by the retrieved chunks; 
(iv) \textit{hallucination (HL)}, the proportion of incorrect 
response claims not entailed by 
the ground-truth answer or any of the 
retrieved chunks;
(v) \textit{relevant noise sensitivity (NS$_{\mathrm{r}}$)}, the 
proportion of incorrect response claims entailed by relevant 
chunks; and (vi) \textit{irrelevant noise sensitivity 	
(NS$_{\mathrm{ir}}$)}, the proportion of incorrect response 
claims entailed by irrelevant chunks. A chunk is 
deemed \textit{relevant} if it entails any claim in the 
ground-truth answer, and \textit{irrelevant otherwise}. 

\noindent\textit{\textbf{Results. } }
Table~\ref{tab:ragchecker_results_full} reports the average 
results across all question-answer pairs. For comparison, we 
also include the  best-performing RAG system 
by~\citet{ru2024}, evaluated on several non-legal datasets.

\begin{table*}[t]
    \centering
    \footnotesize
    \caption{Performance of baseline RAG systems evaluated 
    using \textit{RAGChecker} on two legal 
    datasets. 
    Arrows indicate preferred direction 
    ($\uparrow$ higher is better, $\downarrow$ lower is better). 
   }
    \label{tab:ragchecker_results_full}
    
    \begin{threeparttable}[t]
        
        \begin{tabularx}{\textwidth}{l 
        *{12}{>{\centering\arraybackslash}X}}
            \toprule
            & 
            & \multicolumn{3}{c}{\textbf{Overall}} 
            & \multicolumn{2}{c}{\textbf{Retriever}} 
            & \multicolumn{6}{c}{\textbf{Generator}} \\
            \cmidrule(lr){3-5} \cmidrule(lr){6-7} \cmidrule(lr){8-13}
            
            \textbf{RAG System} & Lang$\dag$
            & \textbf{P$\uparrow$} 
            & \textbf{R$\uparrow$} 
            & \textbf{F1$\uparrow$} 
            & \textbf{CR$\uparrow$} 
            & \textbf{CP$\uparrow$} 
            & \textbf{CU$\uparrow$} 
            & \textbf{NS$_\mathrm{r}$$\downarrow$} 
            & \textbf{NS$_\mathrm{ir}$$\downarrow$}
            & \textbf{\textcolor{black}{HL$\downarrow$}}
            & \textbf{SK$\downarrow$} 
            & \textbf{FT$\uparrow$} \\
            \midrule
            
            BM25+Llama3-8B  & EN    & 45.0 & 62.1 & 48.2 & 
            82.9 & \textbf{76.9} & 69.0 & 48.5 & 2.0 & 
            \textcolor{black}{4.6} & 2.1 & 
            \textbf{93.3} \\
            BM25+Llama3-8B & FR     & 43.6 & 45.5 & 38.3 & 
            65.1 & 28.2 & 46.1 & 24.8 & 13.7 & 
            \textcolor{black}{17.9} & 
            5.9 & 76.1 \\
            
            \midrule
            
            BM25+Mixtral-8x7B  & EN & 49.1 & 68.6 & 52.2 & 
            82.7 & 75.4 & 75.9 & 45.3 & 2.2 & 
            \textcolor{black}{3.4} & 
            3.4 & 93.1 \\
            BM25+Mixtral-8x7B  & FR   & 48.2 & 52.9 & 44.8 & 
            64.7 & 28.3 & 54.1 & 20.9 & 14.5 & 
            \textcolor{black}{16.4} 
            & 
            4.2 & 79.5 \\    
            \midrule
            
            BM25+GPT-4    & EN      & 54.3 & 77.7 & \textbf{60.9} 
            & 82.2 & 76.8 & 83.1 & 40.6 & 2.3 & 
            \textcolor{black}{2.9} 
            & 
            4.7 & 92.4 \\
            BM25+GPT-4      & FR      & 65.6 & 82.7 & 
            \textbf{70.3} & 90.5 & \textbf{34.6} & 85.4 & 27.2 & 
            \textbf{2.5} & \textcolor{black}{4.1} & 1.5 & 93.9 \\
            
            \midrule
            
            BM25+GPT-5    & EN      & 49.6 & \textbf{78.2} & 
56.8 
            & \textbf{86.1} & 70.3 & \textbf{86.6} & 47.5 & 
            \textbf{1.4} & \textcolor{black}{\textbf{1.5}} & 
            \textbf{1.6} & 
            90.9 \\
            BM25+GPT-5     & FR       & 66.4 & \textbf{84.2} & 
            69.9 & \textbf{90.8} & 31.8 & \textbf{86.9} & 28.7 & 3.7 
            & \textcolor{black}{\textbf{1.2}} & \textbf{0.7} & 
            \textbf{98.1} 
            \\
            
            \midrule
            
            E5+Llama3-8B   & EN     & 47.8 & 61.4 & 50.0 & 65.7 
            & 61.9 & 71.6 & 31.1 & 11.6 & \textcolor{black}{9.5} & 
            6.4 & 
            84.1 \\
            E5+Llama3-8B   & FR       & 52.6 & 64.5 & 52.8 & 
80.0 
            & 29.7 & 70.2 & 23.7 & 12.5 & \textcolor{black}{11.2} & 
            3.5 
            & 
            85.4 \\
            
            \midrule
            E5+Mixtral-8x7B & EN    & 49.0 & 56.0 & 47.9 & 66.6 
            & 63.1 & 72.5 & 36.1 & 10.2 & \textcolor{black}{4.7} & 
            3.4 & 
            91.9 \\
            
            E5+Mixtral-8x7B   & FR    & 64.5 & 66.0 & 61.4 & 
80.6 
            & 29.0 & 73.9 & 21.1 & 9.7 & \textcolor{black}{4.7} & 
            1.1 
            & 
            94.2 \\
            \midrule
            E5+GPT-4       & EN     & \textbf{59.8} & 63.2 & 58.3 
& 
            65.4 & 61.7 & 74.5 & 28.8 & 8.4 & 
            \textcolor{black}{3.0} & 
            7.9 & 89.1 \\
            E5+GPT-4      & FR        & \textbf{69.7} & 66.6 & 64.9 
            & 80.3 & 28.5 & 74.7 & \textbf{18.6} & 8.5 & 
            \textcolor{black}{3.3} & 2.7 & 94.0 
            \\
            \midrule
            E5+GPT-5     & EN       & 59.5 & 69.0 & 60.7 & 65.4 & 
            53.7 & 80.6 & \textbf{26.2} & 8.6 & 
            \textcolor{black}{5.6} & 
            8.6 & 85.7 \\
            E5+GPT-5     & FR         & 64.5 & 72.5 & 65.0 & 79.3 
& 
            23.8 & 77.8 & 20.2 & 10.3 & \textcolor{black}{4.9} & 
            3.7 
            & 
            91.4 \\
            \midrule
            \midrule                    
            \citet{ru2024}$\ddag$    &      & 62.0 & 53.0 & 52.7 & 
            83.5 & 61.8 & 60.4 & 28.9 & 3.5 & 
            \textcolor{black}{5.7} & 
            1.4 & 92.9 \\

            \bottomrule
            
        \end{tabularx}
        \begin{tablenotes}
            \item[$\dag$] EN = GDPR and FR = CIVIL 
            \it \item[$\ddag$] best reported results 
            by~\citet{ru2024} on 
            generic domain using E5+GPT-4, obtained using 
            datasets covering only English language queries and 
            corpora. 
        \end{tablenotes}
    \end{threeparttable}
    	\vspace*{-1em}
\end{table*}

Overall, most RAG systems achieved higher P, R, 
and F1 scores on the FR subset than on the EN one.  
Since legal applications prioritize correctness over 
completeness, precision is particularly important: providing 
incorrect legal information can have more severe 
consequences than omitting relevant details, which users may 
still recover through additional search.

Under this criterion, 
E5+GPT4 achieved the best overall performance, marginally 
outperforming E5+GPT5; this confirms the results 
in~\citet{ru2024}. 
Although BM25+GPT5 yielded the highest recall, its lower 
precision, especially on EN questions, makes it less 
attractive for legal use.  
We also remark that 
all systems achieved substantially higher scores on \claims 
than those reported by~\citet{ru2024}.
A likely explanation is the difference in dataset scale: \claims 
contains 394 questions derived from two 
legal documents, whereas the benchmark of \citet{ru2024} 
comprises 4162 questions spanning more than one million 
documents.

The retrieval results reveal that BM25 consistently 
outperforms E5 in both CR and CP across languages. This 
suggests that \textit{dense 
retrievers 
may excel in retrieving semantically related contexts, 
but do not necessarily identify legally relevant ones}. In 
contrast, BM25's lexical matching appears better aligned with 
legal terminology and document structure. This observation 
explains that keyword-based 
retrieval methods remain highly competitive in 
legal applications~\cite{reimers2021}. 

We also observed relatively low CP values for CIVIL despite the 
high CR value achieved on both datasets. A likely reason is that 
CIVIL articles are typically shorter than GDPR articles, causing 
individual chunks to span multiple articles and thus contain 
both relevant and irrelevant content. This result highlights that  
\textit{retrieval effectiveness depends not only on the number 
of retrieved chunks ($k$), but also on their legal relevance and 
granularity.}

The effectiveness of the generation component is reflected in 
the faithfulness (FT) and context utilization (CU) metrics.
GPT-4o and GPT-5 paired with BM25 achieved particularly 
strong results, with FT exceeding 90\% and CU exceeding 
80\% in both languages.
Other models exhibited greater variability. 
For example, Mixtral-8x7B achieved substantially higher FT in 
FR when paired with E5 (94.2\%) than with BM25 
(79.5\%), while it seemed
less affected for EN. Similarly, 
Llama-3.1-8B benefitted from E5 in FR (from 
76.1\% to 85.4\%) but performed worse in EN (from 93.3\% 
to 84.1\%). 
CU results generally improved for these two models paired with 
E5, except for a small drop for Mistral-8x7b for 
EN. 

SK and HL results provide a 
complementary view of the observations 
above. 
GPT-5 achieved the highest SK scores while also maintaining 
the lowest HL rates. 
In general, systems that rely more heavily on retrieved 
evidence (high CU) tend to hallucinate less. 
Conversely, when generators depend more on internal 
knowledge, hallucinations become more frequent.
This pattern is particularly evident for BM25+Llama-3.1-8B 
and BM25+Mixtral-8x7B, whose higher HL rates (17.9 and 
16.4, 
respectively) coincide with lower CU values, especially on 
FR questions.
The markedly higher HL rates in FR are also consistent 
with lower CP and FT scores, suggesting that weaker retrieval 
quality increases the likelihood of hallucinations.

Finally, NS results show that incorrect claims 
arise predominantly from partially relevant chunks rather than 
entirely irrelevant ones. Across all systems, 
NS$_{\mathrm{r}}$ was consistently higher than 
NS$_{\mathrm{ir}}$, indicating that generators are more likely 
to misinterpret or over-generalize information from relevant 
contexts than to fabricate content from unrelated passages.


\subsection{Accuracy of Claim Analysis
in Legal Settings (RQ2)}
\label{sec:rq2_setup}

To assess the reliability of claim analysis in the legal domain, 
we evaluated RefChecker~\cite{hu2024}  on \claims. Consistent 
with its original 
implementation, we used GPT-4-0613 as the enabler 
LLM.

Our evaluation comprises two steps: (i) 
assessing the accuracy of claim extraction from generated 
responses and (ii) assessing the accuracy of claim verification, 
namely determining the entailment relationship between a 
claim and a reference text.

\noindent\textit{\textbf{Evaluation Procedure. } }
During the construction of \claims, 
we 
used RefChecker to extract the claims from 152 answers, 
producing 968 claims in total. 
These claims were manually validated by the legal expert (see 
Appendix~\ref{sec:appendix}), who labeled each claim as valid 
or invalid. 

For claim extraction, we computed \textit{Extraction 
	Accuracy} as the proportion of valid claims among all 
	extracted claims. 
Although this metric primarily reflects precision, rather than 
the ability to identify missed claims, it remains informative for 
assessing the reliability of claim analysis 
as an 
intermediate step in fine-grained RAG evaluation.

For claim verification, 
RefChecker 	assigns one of three labels to each claim: 
\textit{entailment} indicating whether 
	the 
	claim extracted from the answer is entailed in the 
	retrieved 
	contexts, \textit{contradiction} when the claim 
	contradicts the 
	contexts, and \textit{neutral} otherwise. 
To isolate the verification component, we bypassed claim 
extraction and provided RefChecker with the expert-validated 
claims (453 for GDPR and 385 for CIVIL) together with the 
retrieved contexts.

\noindent\textit{\textbf{Results. }}
For claims extraction, RefChecker achieved an accuracy of 
98\% (452/461) for 
GDPR 
(EN) and  98.7\% (385/390) for CIVIL (FR). These results 
indicate that \textit{RefChecker can reliably decompose legal 
responses into valid claims across both EN and FR.}

\begin{table}[t]
	\centering
	\footnotesize
	\setlength{\tabcolsep}{4pt}
	\begin{threeparttable}
		\caption{Claim verification against expert labels. P: 		
		Precision (\%); R.: Recall (\%); F1: F1-score (\%).
		}
		\label{tab:checker_performance}
		\begin{tabularx}{\columnwidth}{@{} l 
		*{4}{>{\centering\arraybackslash}X} 
		*{4}{>{\centering\arraybackslash}X} @{}}
			\toprule
			& \multicolumn{4}{c}{\textbf{GDPR (EN)}} 
			& \multicolumn{4}{c}{\textbf{CIVIL (FR)}} \\
			\cmidrule(lr){2-5} \cmidrule(lr){6-9}
			& $\sum_\text{C}$ & \textbf{P} & \textbf{R} & \textbf{F1}
			& $\sum_\text{C}$ & \textbf{P} & \textbf{R} & 
			\textbf{F1} 
			\\
			\midrule
			$\epsilon$   & 295 & 84.6 & 73.4 & 78.6 & 284 & 89.3 
			& 82.4 
			& 85.7 \\
			$\eta$ & 151 & 68.1 & 72.2 & 70.1 & 77  & 53.3 & 62.3 
			& 
			57.5 \\
 			$\zeta$ & 7   & 2.6  & 14.3 & 4.4  & 24  & 24.2 & 33.3 
 			& 
			28.1 \\
			
			\bottomrule
		\end{tabularx}
			    \begin{tablenotes} 
			\small
			\it \item $\sum_\text{C}$ is the number of claims 
			per class in \claims according to the expert's manual 
			analysis; \\
			$\epsilon$, 
			$\eta$, $\zeta$ refer to 
			entailment, neutral, 
			and 
			contradiction, respectively.
			\end{tablenotes}
	\end{threeparttable}
	\vspace*{-1.5em}
\end{table}

Table~\ref{tab:checker_performance} reports the performance 
of RefChecker's verification labels against the expert 
annotations. 
The results reveal a clear weakness in identifying contradictory 
claims. The average F1 score for the \textit{contradiction} 
class is only 4.4\% on GDPR and 28.1\% on CIVIL. The 
corresponding precision and recall values, particularly on 
GDPR, indicate that RefChecker both misses many genuine 
contradictions and incorrectly labels non-contradictory claims 
as contradictions. As a result, errors in claim verification may 
propagate to the downstream evaluation of RAG systems.

Although contradictory claims are relatively rare in our dataset, 
the results suggest that contradiction detection remains a 
challenging problem in legal settings. \textit{Overall, while 
RefChecker performs very well at claim extraction, its 
verification component, especially for contradiction detection, 
requires further improvement for reliable legal-domain 
evaluation.}


\subsection{Threats to Validity and Limitations}
\label{sec:threats-to-validity}

\noindent\textit{\textbf{Threats to Validity.}}
Some question categories like 
false-premise are underrepresented (5.4\% in GDPR; 3.0\%  in 
CIVIL), limiting  the strength of conclusions for these cases. In 
addition, the dataset includes only two languages and one 
document per language, limiting linguistic and document-level 
diversity.
All QA pairs were generated using GPT-4, which is 
also one of the evaluated generators and may therefore 
benefit from generation bias.

The manual validation was performed by a single legal expert. 
Although annotation quality was supported through a pilot 
phase, detailed guidelines, and iterative feedback, the 
absence of multiple annotators prevents the computation of 
inter-annotator agreement metrics. 

\noindent\textit{\textbf{Limitations. }} As a synthetic dataset, 
\claims may not 
fully capture the diversity and complexity of real-world legal 
queries, which may limit the generalizability of our findings. 
Moreover, although the dataset is suitable for benchmarking 
legal RAG systems, it should not be used for legal advice 
without expert review.

\section{Conclusion} 
\label{sec:conclusion}
This paper has introduced \claims, a bilingual, bi-jurisdictional 
benchmark for fine-grained legal RAG evaluation. The 
benchmark contains 317 expert-validated QA pairs, spanning 
diverse question categories and user personas, together with 
968 manually validated claims. Using \claims, we have benchmarked
eight state-of-the-art RAG systems and identified 
domain-specific weaknesses in both retrieval and generation
that are not captured by existing legal RAG benchmarks. We 
also showed that current claim-level evaluation frameworks, 
while effective in 
general-domain settings, exhibit important limitations in legal 
contexts, particularly for contradiction detection. 
Future work includes a deeper investigation of hallucinations in 
legal RAG systems, expanding underrepresented question 
categories, and enriching \claims with additional 
expert-authored QA pairs and claims.

\bibliographystyle{plainnat}
\bibliography{main}

@article{hu2024,
  title={{RefChecker}: Reference-based fine-grained 
  hallucination checker and benchmark for large language 
  models},
  author={Hu, Xiangkun and Ru, Dongyu and Qiu, Lin and Guo, Qipeng and Zhang, Tianhang and Xu, Yang and Luo, Yun and Liu, Pengfei and Zhang, Yue and Zhang, Zheng},
  journal={arXiv preprint arXiv:2405.14486},
  year={2024},
  url={https://doi.org/10.48550/arXiv.2405.14486}
}

@article{ru2024,
  title={{RAGChecker}: A fine-grained framework for 
  diagnosing 
  retrieval-augmented generation},
  author={Ru, Dongyu and Qiu, Lin and Hu, Xiangkun and Zhang, Tianhang and Shi, Peng and Chang, Shuaichen and Jiayang, Cheng and Wang, Cunxiang and Sun, Shichao and Li, Huanyu and others},
  journal={Advances in Neural Information Processing Systems},
  volume={37},
  pages={21999--22027},
  year={2024},
  url={https://doi.org/10.52202/079017-0692}
}

@inproceedings{es2024,
  title={{RAGAs}: Automated evaluation of retrieval augmented 
  generation},
  author={Es, Shahul and James, Jithin and Anke, Luis 
  Espinosa and Schockaert, Steven},
  booktitle={Proceedings of the 18th Conference of the European Chapter of the Association for Computational Linguistics: System Demonstrations},
  pages={150--158},
  year={2024},
  url={https://doi.org/10.18653/v1/2024.eacl-demo.16}
}

@inproceedings{li2025,
  title={{RAG-Zeval: Enhancing RAG Responses Evaluator 
  through End-to-End Reasoning and Ranking-Based 
  Reinforcement Learning}},
  author={Li, Kun and Li, Yunxiang and Zhang, Tianhua and Luo, Hongyin and Wu, Xixin and Glass, James and Meng, Helen},
  booktitle={Proceedings of the 2025 Conference on Empirical Methods in Natural Language Processing},
  pages={24936--24954},
  year={2025},
  url={https://doi.org/10.18653/v1/2025.emnlp-main.1267}
}

@article{magesh2025,
  title={Hallucination-Free? Assessing the Reliability of Leading 
  {AI} Legal Research Tools},
  author={Magesh, Varun and Surani, Faiz and Dahl, Matthew and Suzgun, Mirac and Manning, Christopher D and Ho, Daniel E},
  journal={Journal of Empirical Legal Studies},
  volume={22},
  number={2},
  pages={216--242},
  year={2025},
  publisher={Wiley Online Library},
  url={https://doi.org/10.1111/jels.12413}
}

@article{dahl2024,
	title={Large legal fictions: Profiling legal hallucinations in large language models},
	author={Dahl, Matthew and Magesh, Varun and Suzgun, Mirac and Ho, Daniel E},
	journal={Journal of Legal Analysis},
	volume={16},
	number={1},
	pages={64--93},
	year={2024},
	publisher={Oxford University Press UK},
  url={https://doi.org/10.1093/jla/laae003}
}

@article{pipitone2024,
  title={{LegalBench-RAG}: A benchmark for 
  retrieval-augmented generation in the legal domain},
  author={Pipitone, Nicholas and Alami, Ghita Houir},
  journal={arXiv preprint arXiv:2408.10343},
  year={2024},
  url={https://doi.org/10.48550/arXiv.2408.10343}
}

@book{alammar2024,
	title={Hands-on large language models: language understanding and generation},
	author={Alammar, Jay and Grootendorst, Maarten},
	year={2024},
	publisher={" O'Reilly Media, Inc."}
}

@article{gokhan2024,
  title={{RIRAG}: Regulatory Information Retrieval and Answer 
  Generation},
  author={Gokhan, Tuba and Wang, Kexin and Gurevych, Iryna and Briscoe, Ted},
  journal={arXiv preprint arXiv:2409.05677},
  year={2024},
  url={https://doi.org/10.48550/arXiv.2409.05677}
}

@article{mallick2024,
	title={{Generative AI} in the Law},
	author={Mallick, Samuel},
	journal={the Law (February 10, 2024)},
	volume={42},
	year={2024},
  url={https://doi.org/10.2139/ssrn.5040429}
}

@misc{edwards2025,
	author       = {Ben Edwards},
	title        = {Number of legal professionals using {Gen AI} 
	jumps sharply over past year, study shows},
	year         = {2025},
	month        = {April 17},
	howpublished = {\href{https://www.globallegalpost.com/news/number-of-legal-professionals-using-gen-ai-jumps-sharply-over-past-year-study-shows-1273491086}{number-of-legal-professionals-using-gen-ai}},
	note         = {Accessed: 2026-01-04},
	publisher    = {The Global Legal Post},
}

@misc{weiser2023,
	author       = {Benjamin Weiser},
	title        = {{‘I apologise for the confusion earlier’: Here’s 
	what happens when your lawyer uses ChatGPT'}},
	year         = {2023},
	month        = {May 28},
	howpublished = {\href{https://www.irishtimes.com/world/us/2023/05/28/i-apologize-for-the-confusion-earlier-heres-what-happens-when-your-lawyer-uses-chatgpt/}{heres-what-happens-when-your-lawyer-uses-chatgpt}},
	note         = {Accessed: 2026-01-04},
	publisher    = {The Irish Times}
}

@article{keisha2025,
	title={All for law and law for all: Adaptive {RAG} Pipeline for 
	Legal Research},
	author={Keisha, Figarri and Singh, Prince and Fernandes, Dion and Manivannan, Aravindh and Wicaksono, Ilham and Ahmad, Faisal and Rim, Wiem Ben and others},
	journal={arXiv preprint arXiv:2508.13107},
	year={2025},
  url={https://doi.org/10.48550/arXiv.2508.13107}
}

@article{lewis2020,
	title={Retrieval-augmented generation for knowledge-intensive nlp tasks},
	author={Lewis, Patrick and Perez, Ethan and Piktus, Aleksandra and Petroni, Fabio and Karpukhin, Vladimir and Goyal, Naman and K{\"u}ttler, Heinrich and Lewis, Mike and Yih, Wen-tau and Rockt{\"a}schel, Tim and others},
	journal={Advances in neural information processing systems},
	volume={33},
	pages={9459--9474},
	year={2020},
  url={https://doi.org/10.48550/arXiv.2005.11401}
}

@misc{esquire2025,
author       = {{Esquire Deposition Solutions}},
title        = {Federal Court Turns Up the Heat on Attorneys 
Using {ChatGPT} for Research},
year         = {2025},
month        = {August 13},
howpublished = {\href{https://www.esquiresolutions.com/federal-court-turns-up-the-heat-on-attorneys-using-chatgpt-for-research/}{federal-court-turns-up-the-heat-on-attorneys}},
note         = {Accessed: 2026-01-04},
publisher    = {Esquire Deposition Solutions},
}

@inproceedings{kalai2024,
title={Calibrated language models must hallucinate},
author={Kalai, Adam Tauman and Vempala, Santosh S},
booktitle={Proceedings of the 56th Annual ACM Symposium on Theory of Computing},
pages={160--171},
year={2024},
url={https://doi.org/10.1145/3618260.3649777}
}

@inproceedings{guha2023,
	title={{LegalBench}: A collaboratively built benchmark for 
	measuring legal reasoning in large language models},
	author={Guha, Neel and Nyarko, Julian and Ho, Daniel and
 others},
	booktitle={Proceedings of the 37th Conference on Neural
                  Information Processing Systems - Datasets and 
                  Benchmarks Track},
	pages={44123--44279},
	year={2023}
}

@inproceedings{niklaus2023,
	title={{LEXTREME}: A Multi-Lingual and Multi-Task Benchmark for the Legal Domain},
	author={Niklaus, Joel and Matoshi, Veton and Rani, Pooja 
	and Galassi, Andrea and St{\"u}rmer, Matthias and Chalkidis, 
	Ilias},
	booktitle={Findings of the Association for Computational Linguistics: EMNLP 2023},
	pages={3016--3054},
	year={2023},
  url={https://doi.org/10.18653/v1/2023.findings-emnlp.200}
}

@inproceedings{chalkidis2022,
  title={{LexGLUE}: A benchmark dataset for legal language 
  understanding in English},
  author={Chalkidis, Ilias and Jana, Abhik and Hartung, Dirk and Bommarito, Michael and Androutsopoulos, Ion and Katz, Daniel and Aletras, Nikolaos},
  booktitle={Proceedings of the 60th Annual Meeting of the Association for Computational Linguistics (Volume 1: Long Papers)},
  pages={4310--4330},
  year={2022},
  url={https://doi.org/10.18653/v1/2022.acl-long.297}
}

@inproceedings{hou2025,
  title={{CLERC}: A dataset for US legal case retrieval and 
  retrieval-augmented analysis generation},
  author={Hou, Abe Bohan and Weller, Orion and Qin, Guanghui and Yang, Eugene and Lawrie, Dawn and Holzenberger, Nils and Blair-Stanek, Andrew and Van Durme, Benjamin},
  booktitle={Findings of the Association for Computational Linguistics: NAACL 2025},
  pages={7913--7928},
  year={2025},
  url={https://doi.org/10.18653/v1/2025.findings-naacl.441}
}

@inproceedings{wiratunga2024,
  title={{CBR-RAG}: case-based reasoning for retrieval 
  augmented generation in LLMs for legal question answering},
  author={Wiratunga, Nirmalie and Abeyratne, Ramitha and Jayawardena, Lasal and Martin, Kyle and Massie, Stewart and Nkisi-Orji, Ikechukwu and Weerasinghe, Ruvan and Liret, Anne and Fleisch, Bruno},
  booktitle={International Conference on Case-Based Reasoning},
  pages={445--460},
  year={2024},
  organization={Springer},
  url={https://doi.org/10.1007/978-3-031-63646-2_29}
}

@inproceedings{lee2025koblex,
  title={{KoBLEX}: Open Legal Question Answering with 
  Multi-hop Reasoning},
  author={Lee, Jihyung and Kim, Daehui and Hwang, Seonjeong and Kim, Hyounghun and Lee, Gary},
  booktitle={Proceedings of the 2025 Conference on Empirical Methods in Natural Language Processing},
  pages={4019--4053},
  year={2025},
  url={https://doi.org/10.18653/v1/2025.emnlp-main.200}
}

@inproceedings{louis2024,
  title={Interpretable long-form legal question answering with retrieval-augmented large language models},
  author={Louis, Antoine and van Dijck, Gijs and Spanakis, Gerasimos},
  booktitle={Proceedings of the AAAI Conference on Artificial Intelligence},
  xxvolume={38},
  xxnumber={20},
  pages={22266--22275},
  year={2024},
  url={https://doi.org/10.1609/aaai.v38i20.30232}
}

@inproceedings{min2023,
  title={{FActScore}: Fine-grained atomic evaluation of factual 
  precision in long form text generation},
  author={Min, Sewon and Krishna, Kalpesh and Lyu, Xinxi and Lewis, Mike and Yih, Wen-tau and Koh, Pang and Iyyer, Mohit and Zettlemoyer, Luke and Hajishirzi, Hannaneh},
  booktitle={Proceedings of the 2023 Conference on Empirical 
  Methods in Natural Language Processing},
  pages={12076--12100},
  year={2023},
  url={https://doi.org/10.18653/v1/2023.emnlp-main.741}
}

@inproceedings{scire2024,
  title={{FENICE}: Factuality Evaluation of summarization based 
  on Natural language Inference and Claim Extraction},
  author={Scir{\`e}, Alessandro and Ghonim, Karim and Navigli, Roberto},
  booktitle={Findings of the Association for Computational Linguistics ACL 2024},
  pages={14148--14161},
  year={2024},
  url={https://doi.org/10.18653/v1/2024.findings-acl.841}
}

@article{wei2024,
  title={Long-form factuality in large language models},
  author={Wei, Jerry and Yang, Chengrun and Song, Xinying and Lu, Yifeng and Hu, Nathan and Huang, Jie and Tran, Dustin and Peng, Daiyi and Liu, Ruibo and Huang, Da and others},
  journal={Advances in Neural Information Processing Systems},
  volume={37},
  pages={80756--80827},
  year={2024},
  url = {https://doi.org/10.52202/079017-2567}
}

@article{robertson2009,
author = {Robertson, Stephen and Zaragoza, Hugo},
title = {The Probabilistic Relevance Framework: {BM25} and 
Beyond},
year = {2009},
issue_date = {April 2009},
publisher = {Now Publishers Inc.},
address = {Hanover, MA, USA},
volume = {3},
number = {4},
issn = {1554-0669},
url = {https://doi.org/10.1561/1500000019},
xxdoi = {10.1561/1500000019},
journal = {Found. Trends Inf. Retr.},
month = apr,
pages = {333–389},
numpages = {57}
}

@article{wang2023,
  title={Improving Text Embeddings with Large Language Models},
  author={Wang, Liang and Yang, Nan and Huang, Xiaolong and Yang, Linjun and Majumder, Rangan and Wei, Furu},
  journal={arXiv preprint arXiv:2401.00368},
  year={2023},
  url={https://doi.org/10.48550/arXiv.2401.00368}
}

@article{jiang2024,
  title={Mixtral of experts},
  author={Jiang, Albert Q and Sablayrolles, Alexandre and Roux, Antoine and  others},
  journal={arXiv preprint arXiv:2401.04088},
  year={2024},
  url={https://doi.org/10.48550/arXiv.2401.04088}
}

@article{grattafiori2024,
  title={The {Llama} 3 herd of models},
  author={Grattafiori, Aaron and Dubey, Abhimanyu and Jauhri, Abhinav and others},
  journal={arXiv preprint arXiv:2407.21783},
  year={2024},
  url={https://doi.org/10.48550/arXiv.2407.21783}
}

@inproceedings{reimers2021,
	title = "The Curse of Dense Low-Dimensional Information 
	Retrieval for Large Index Sizes",
	author = "Reimers, Nils  and Gurevych, Iryna",
	booktitle = "Proceedings of the 59th Annual Meeting of the 
	Association for Computational Linguistics and the 11th 
	International Joint Conference on Natural Language 
	Processing (Volume 2: Short Papers)",
	year = "2021",
	publisher = "Association for Computational Linguistics",
	  url={https://doi.org/10.18653/v1/2021.acl-short.77},
	pages = "605--611",
}

@misc{EUGDPR,
	author = {{The European Parliament and the Council of the 
	European Union}},
	title = {Regulation (EU) 2016/679 of the European Parliament 
	and of the Council of 27 April 2016 on the protection of 
	natural persons with regard to the processing of personal 
	data and on the free movement of such data, and repealing 
	Directive 95/46/EC 
	({General Data Protection Regulation})},
	number = {119},
	year = {2016},
	month = {05},
	pages = {1-88},
	url = {https://eur-lex.europa.eu/eli/reg/2016/679/oj}
}

@misc{CIVIL,
	title     = {National Civil Code (consolidated version)},
	author    = {{Grand-Duch{\'{e}} de Luxembourg}},
	publisher = {Journal officiel du Grand-Duch{\'{e}} de 
	Luxembourg (Legilux)},
	url       = 
	{https://legilux.public.lu/},
	urldate   = {2026-01-19},
	year = {2025},
}

@inproceedings{manakul2023,
  title={{SelfCheckGPT}: Zero-resource black-box hallucination 
  detection for generative large language models},
  author={Manakul, Potsawee and Liusie, Adian and Gales, Mark},
  booktitle={Proceedings of the 2023 conference on empirical methods in natural language processing},
  pages={9004--9017},
  year={2023},
  url={https://doi.org/10.18653/v1/2023.emnlp-main.557}
}

@inproceedings{wang2025,
  title={OpenFactCheck: Building, benchmarking customized fact-checking systems and evaluating the factuality of claims and LLMs},
  author={Wang, Yuxia and Wang, Minghan and Iqbal, Hasan and Georgiev, Georgi N and Geng, Jiahui and Gurevych, Iryna and Nakov, Preslav},
  booktitle={Proceedings of the 31st international conference on computational linguistics},
  pages={11399--11421},
  year={2025},
  url={https://aclanthology.org/2025.coling-main.755/}
}

@article{metropolitansky2025,
  title={Veritrail: Closed-domain hallucination detection with traceability},
  author={Metropolitansky, Dasha and Larson, Jonathan},
  journal={arXiv preprint arXiv:2505.21786},
  year={2025},
  url={https://doi.org/10.48550/arXiv.2505.21786}
}

@inproceedings{metropolitansky2025_clamify,
  title={Towards effective extraction and evaluation of factual claims},
  author={Metropolitansky, Dasha and Larson, Jonathan},
  booktitle={Proceedings of the 63rd Annual Meeting of the Association for Computational Linguistics (Volume 1: Long Papers)},
  pages={6996--7045},
  year={2025},
  url={https://doi.org/10.18653/v1/2025.acl-long.348}
}

@misc{trulens,
	title = {{The RAG Triad}},
	howpublished = 
	{\url{https://www.trulens.org/getting_started/core_concepts/rag_triad/}},
	note = {Accessed: 2026-04-28},
	author    = {Joe Ferrara and {Ethan-Tonic} and Oguzhan 
	Mete Ozturk},
	year = {2024},
}

@article{hurst2024,
	title={{GPT}-4o system card},
	author={Hurst, Aaron and Lerer, Adam and Goucher, Adam P 
	and Perelman, Adam and Ramesh, Aditya and Clark, Aidan and 
	Ostrow, AJ and Welihinda, Akila and Hayes, Alan and Radford, 
	Alec and others},
	journal={arXiv preprint arXiv:2410.21276},
	year={2024},
  url={https://doi.org/10.48550/arXiv.2410.21276}
}

@article{singh2025,
	title={{OpenAI} {GPT}-5 system card},
	author={Singh, Aaditya and Fry, Adam and Perelman, Adam and 
	Tart, Adam and Ganesh, Adi and El-Kishky, Ahmed and 
	McLaughlin, Aidan and Low, Aiden and Ostrow, AJ and 
	Ananthram, Akhila and others},
	journal={arXiv preprint arXiv:2601.03267},
	year={2025},
  url={https://doi.org/10.48550/arXiv.2601.03267}
}

\newpage
\appendix
\onecolumn

\end{document}